\documentclass[letterpaper]{article} 
\usepackage{aaai25}  
\usepackage{times}  
\usepackage{helvet}  
\usepackage{courier}  
\usepackage[hyphens]{url}  
\usepackage{graphicx} 
\usepackage{natbib}  
\usepackage{caption} 
\usepackage{algorithm}
\usepackage{algorithmic}
\usepackage{xcolor}

\usepackage{newfloat}
\usepackage{listings}

\usepackage{subfigure}
\usepackage{booktabs}
\usepackage{caption}
\usepackage{subcaption}

\usepackage{amsmath}
\usepackage{amssymb}
\usepackage{mathtools}
\usepackage{amsthm}
\usepackage{multirow}
\usepackage{colortbl}
\usepackage{amssymb}

\DeclareCaptionStyle{ruled}{labelfont=normalfont,labelsep=colon,strut=off} 
\floatstyle{ruled}
\newfloat{listing}{tb}{lst}{}
\floatname{listing}{Listing}
\title{WarmFed: Federated Learning with Warm-Start for Globalization and Personalization Via Personalized Diffusion Models}
\author{
    Tao Feng$^{1\star}$\quad 
    Jie Zhang$^{2\star}$\quad  
    Xiangjian Li$^{1}$\quad 
    Rong Huang$^{1}$\quad 
    Huashan Liu$^{1\dagger}$\quad
    Zhijie Wang$^{1}$\quad
}
    
\affiliations{
    \textsuperscript{\rm 1}College of Information Science and Technology, Donghua University, Shanghai, China\\
    \textsuperscript{\rm 2}Department of Computer Science, ETH Zurich, Zurich, Switzerland\\

}

\usepackage{bibentry}

\begin{document}

\maketitle

\begin{abstract}
Federated Learning (FL) stands as a prominent distributed learning paradigm among multiple clients to achieve a unified global model without privacy leakage. In contrast to FL, Personalized federated learning aims at serving for each client in achieving persoanlized model. However, previous FL frameworks have grappled with a dilemma: the choice between developing a singular global model at the server to bolster globalization or nurturing personalized model at the client to accommodate personalization. Instead of making trade-offs, this paper commences its discourse from the pre-trained initialization, obtaining resilient global information and facilitating the development of both global and personalized models.  Specifically, we propose a novel method called \textbf{WarmFed} to achieve this. WarmFed customizes Warm-start through personalized diffusion models, which are generated by local efficient fine-tunining (LoRA). Building upon the Warm-Start, we advance a server-side fine-tuning strategy to derive the global model, and propose a dynamic self-distillation (DSD) to procure more resilient personalized models simultaneously. Comprehensive experiments underscore the substantial gains of our approach across both global and personalized models, achieved within just one-shot and five communication(s).
\end{abstract}

\maketitle

\section{Introduction}
The widespread availability of online training data has significantly empowered the capabilities of deep learning models. Now, practical considerations in certain companies or organizations, such as banking, medical and manufacturers institutions, necessitate the strict confidentiality of client data, rendering centralized learning impractical. Federated learning~\citep{massoulie2007randomized,konevcny2015federated} then emerges as a collaborative training strategy without compromising the privacy of decentralized clients in this context.   

The standard setting for FL~\citep{konevcny2016federated}, is to train a single global model that performs well on the globalized data of clients, which we call it GFL here. GFL coordinates local training and server aggregation to derive global information (\textit{e.g.}, global model parameters), thereby engendering a single global model through iterative communications. Recently, Personalized Federated learning (PFL) \citep{smith2017federated,tan2022towards} has been gained attention for its great performance on clients, especially in hospitals and different individuals which owns different preferences, behavior patterns. PFL predominantly emphasizes local training for achieving personalized model of each client, which leverages the server-aggregated global information to address the data shortage problem. Albeit the disparate objectives of GFL and PFL, their optimization mechanisms share similarity in leveraging global information. In other word, they both utilize global information to dismantle data silos, culminating in the emergence of global or personalized models.

Inspired by transfer learning, recent studies \citep{nguyen2022begin,chen2022importance,tan2022federated,zhang2023federated,learninggpt} have directed attention towards activating FL with pre-trained start instead of random start \citep{li2020federated,li2021fedbn,fallah2020personalized}. Chen et all. \citep{chen2022importance} demonstrate that pre-trained start of FL does not inherently address the model drift stemming from the non-IID data of clients, the efficacy of which should be attributed to the large global aggregation gain. In other words, pre-trained start provide robust global information for FL. \textit{In this paper, we principally focuses on the start of FL, delving into the robust global information for both global and personalized models.}

With the capacity to generate high-fidelity images demonstrated by fundamental generative models \citep{rombach2022high, ramesh2022hierarchical, nichol2021glide}, several FL frameworks \citep{zhang2023federated,learninggpt} opt for adapting dynamic pre-trained model, which is trained with synthetic data that exhibits the same category of the private data. They generate synthetic data in the server through off-the-shelf stable diffusion model \citep{rombach2022high} with corresponding uploaded prompts by clients through BLIP-v2 \citep{li2023blip} in FGL \citep{zhang2023federated} or GPT-3 \citep{brown2020language} in GPFL \citep{learninggpt}. Unlike traditional pre-trained models with fixed start, the utilization of synthetic data through prompts stands out as a more effective strategy for crafting personalized initialization schemes to achieve commendable results, they may not adapt to intricate datasets, especially fine-grained and medical datasets. The root cause of this issue lies in the inherent difficulty of accurately encapsulating the stylistic information of private datasets within the text descriptions (i.e., prompts) alone. Besides, the prompt itself constitutes a detailed description of each image, which poses a potential risk of privacy leakage. Consequently, we pose the following consideration:

\textit{How to establish an efficient and privacy-protecting mechanism for the transmission of personal information, thus generating a warm-start that is customized for clients?}

To achieve this goal, we present a method, named \textbf{WarmFed}. Specifically, We obtain warm-start by generating synthetic data through personalized diffusion models, which are fine-tuned locally using lightweight matrix parameters via LoRA \citep{hu2021lora}. Based on warm-start, we realize globalization and personalization, respectively. (1) For globalization, we utilize the synthetic data with global information to both fine-tune the local and aggregated models. (2) For personalization, we propose dynamic self-distillation strategy, which attains the effective personalized models by selecting personalized knowledge dynamically and distilling it to the global model. The proposed WarmFed exhibits multifold merits compared to prior arts. First, personalized diffusion model can procure synthetic data that aligns more closely with client information, thereby facilitating adaptability to more intricate data. Secondly, each parameter matrix is merely 2M in size, rendering low transmission cost relatively. Third, the uploaded parameter matrix does not include any available cotent that could easily violate private data, which affords clients robust privacy protection. 

Our contributions can be summarized as follows:
\begin{itemize}
    \item We explore both global and personalized FL through pre-trained start. To achieve this objective, we propose an innovative technique termed WarmFed, which establishes customized Warm-Start with low transmission cost and high privacy.
    \item We propose the server-side fine-tuning and dynamic self-distillation strategies to further achieve better globalization and personalization based on Warm-Start.
    \item WarmFed gives great performance in only one-shot and five communication(s).
\end{itemize}

\section{Related Works}
\paragraph{Federated Learning}
In recent years, considerable researches has been devoted to advancing FL, which focuses on dismantling data silos without disclosing client privacy. FedAvg \citep{konevcny2016federated} serves as a milestone, endeavors to train a unified global model fitted global distribution. However, the inherent non-IID client data in real word incurs significant deviation. To address this challenge, \citep{acar2021federated}, \citep{li2021model}, \citep{zhou2023fedfa} target on either client side or server side to mitigate client drift. Nevertheless, as the derived global model is designed to fit the “average client”, its efficacy at each client diminishes a lot. Therefore, \citep{li2021fedbn}, \citep{sun2021partialfed} leverage global information to train personalized models that fit the distribution of each client. FedRoD \citep{chen2021bridging}, employs balanced softmax (BSM) \citep{ren2020balanced} and personal head to obtain both global and personal model. Nevertheless,  FedRoD is limmited in other settings, like feature shift. In contrast to these methods, our approach initiates from initialization of global model, concurrently generating global and personalized models in most settings.

\paragraph{Diffusion Model}
Diffusion Models (DM) \citep{rombach2022high,ramesh2022hierarchical,nichol2021glide,saharia2022photorealistic} have gained great attention owing to its formidable capacity for generating high-quality images. Recently, In order to satisfy  personalized  generation in higher subject fidelity, several works fine-tune text-to-image diffusion models given few subject images. Textual Inversion \citep{gal2022image} learns to associate the visual concepts of given images with input text embedding. DreamBooth \citep{ruiz2023dreambooth} proposes a unique identifier linked to user-specific data to optimize the parameters of the entire Text-Image model. LoRa \citep{hu2021lora} and StyleDrop \citep{sohn2023styledrop} have successfully realized personalization by compact weight spaces instead of whole weights. Given the stringent constraints posed by FL, we choose LoRA, a parameter efficient fine-tuning strategy, as our personalized fine-tuning at clients.

\paragraph{Federated Learning with Client Knowledge}
To mitigate the challenge of client drift, recent studies entail the transmission of information enriched with client knowledge to the server. For example, KT-PFL \citep{zhang2021parameterized} strengthen the cooperation among clients with knowledge coefficient matrix through soft predictions generated by public data. However, the predictions of public data are irrelevant with private data frequently, which may not represent local knowledge. Chen et al \citep{chen2023federated}. make customization for local style of each client and send them to the server.  Hu et al. \citep{hu2022fedsynth,xiong2023feddm,pi2023dynafed} utilize dataset distillation \citep{zhao2020dataset,zhao2023dataset} to generate distilled dataset containing intensive information of client data in the clients and transform to the server. Nevertheless, the effect of dataset distillation relies on network architecture to a great extent. And it also faces significant limitations in distilling high-resolution images. To exploit the capabilities of large-scale models in FL, Zhang et al. \citep{zhang2023federated,learninggpt} send the local abundant data-related prompts generated by BLIP-v2 or GPT-3 to the server and untilize Stable Diffusion \citep{rombach2022high} to generate synthetic data with high fidelity. Nevertheless, the prompts fail to express the distinctive data and may cause privacy leakage. In contrast to the methods outlined above, we ensures efficient and accurate transfer of client knowledge by transmitting generative model parameters within a low-dimensional space, while preserving client privacy.

\section{Empirical Study}\label{sec:empirical}
The diverse initialized starts in FL exhibit various degrees of influence on global information, consequently impacting the performance of both global and personalized models, respectively. This section explores both the global and personalized performance under different starts of FL through empirical studies.

\begin{figure}[t]
    \centering
    \centering
    \includegraphics[width=1\columnwidth]{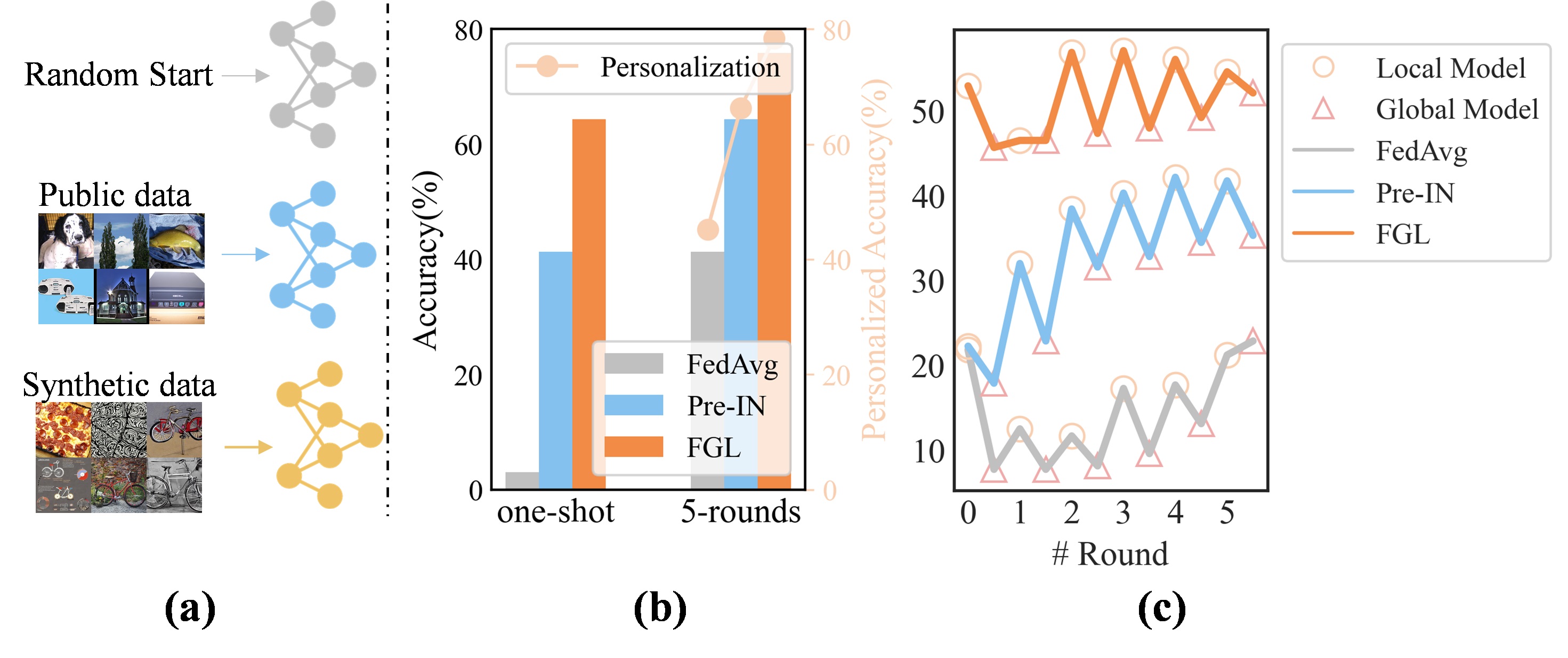}
    \vspace{-0.8cm}
    \caption{\small \textbf{(a)} The performance of personalized model. \textbf{(b)} The details on one client personalization.}
    \label{emp_1}
    \vspace{-0.6cm}
\end{figure}

\textbf{Globalization on different starts.} To investigate the global performance of diverse start in FL, we conduct an experimental study on two kinds of initialized starts: random start (FedAvg \citep{konevcny2016federated}) and pre-trained starts \citep{nguyen2022begin,zhang2023federated} as illustrated in Figure~\ref{emp_1}a. We choose DomianNet and ResNet18 as in FGL \citep{zhang2023federated}. In the pre-trained start, the global model is pre-trained using synthetic data and public data, respectively. For synthetic data, we adopt FGL \citep{zhang2023federated}, where synthetic data is generated through detailed prompts. For public data, we pre-train the global model with a subset of ImageNet1k (Pre-IN) \citep{nguyen2022begin} to ensure a fair comparison, which keeps consistent with the volume of synthetic data above.  As illustrated in Figure~\ref{emp_1}b, the pre-trained start (Pre-IN and FGL) surpasses FedAvg obviously in one-shot and 5 round(s).  Meanwhile, the effect of Pre-IN is inferior to FGL with better start.  


\textbf{Personalization on different starts.} We argue that client personalized capability hold significant potential in local models (the model uploaded to the server after being fine-tuned locally) when using pre-trained initialized start. To verify this, we investigate the personalization of both the global and the local model under different situations.

As shown in Figure~\ref{emp_1}b, we observe that the local model initialized through pre-trained start exhibit high personalization. To further expore the relationship between the generalization of the global model and the personalization of the local model under pre-trained start, we present the personalized performance of global and local model under different rounds in Figure~\ref{emp_1}c. It is evident that the personalization of both local and global models are relatively stable under pre-trained start, with the local model exhibiting higer personalization than the global model steadily.


The empirical studies show that a well start of FL provide the construction of a resilient global model, with local models consistently showing effective and stable personalization. Hence, this raises a critical challenge of how to customize stronger start for FL.

\begin{figure*}[t] 
    \centering
    \includegraphics[width=0.85\textwidth]{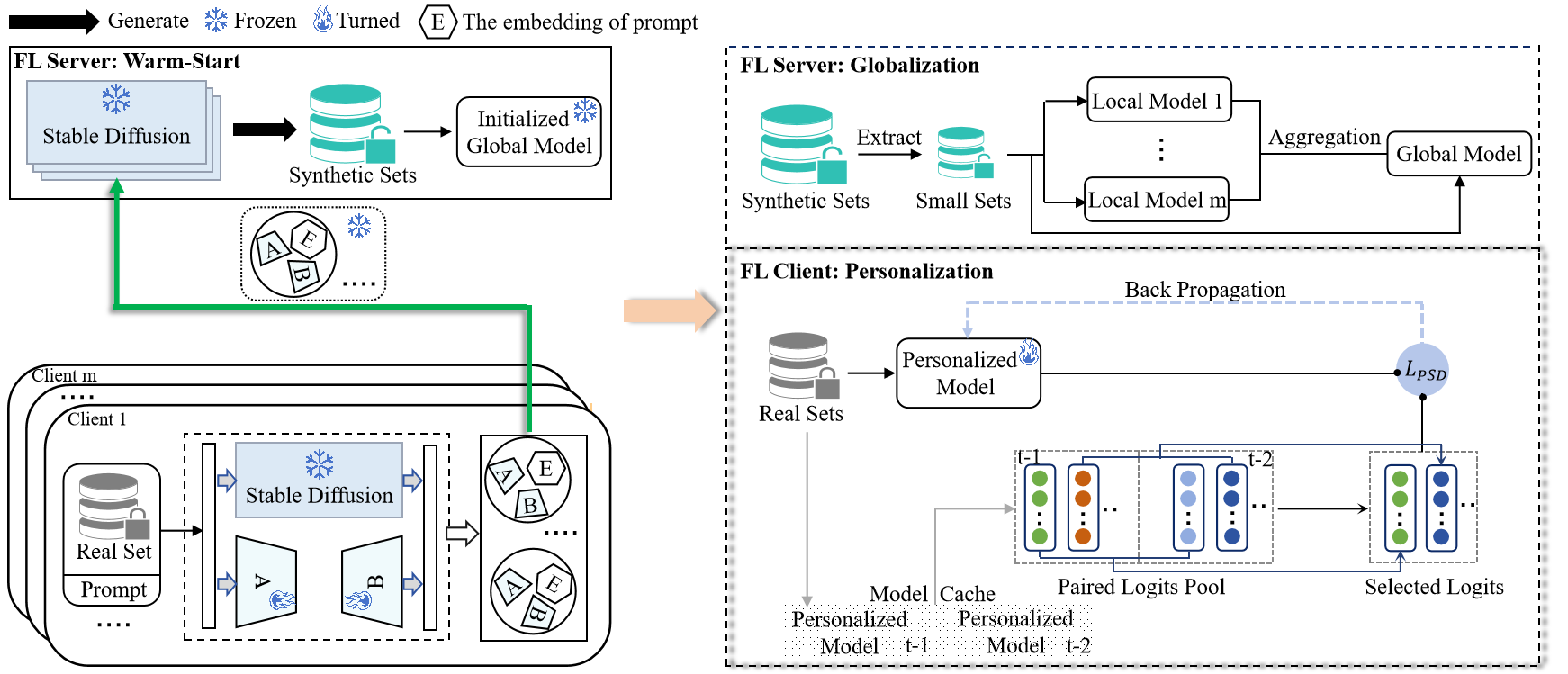}%
    \caption{Pipeline of WarmFed, which consists of three stages: (1)For warm-start stage, we fine-tune SD through LoRA and send the fine-tuned parameter matrixes to the server for warm-start. (2)For globalization stage, we fine-tune the local models and aggregated model with synthetic data to achieve final global model. (3)For personalization stage, the personlized model is obtained by dynamic self-distillation, which is employed to select personlized knowledge for self-distillation.}
    \label{fig:example-image}
    \vspace{-0.3cm}
\end{figure*}

\section{Methodology}
In this section, we propose an efficient federated learning framework named WarmFed, which aims at generating robost start of FL (Warm-Start). Based on this, we consider the globalization and personalization when the data on each client are non-IID. The detailed process is  presented in Figure~\ref{fig:example-image}, and the whole training Algorithm of WarmFed is shown in \textbf{Appendix}. 

\subsection{Warm-Start for Federated Learning}\label{initial_global}

In order to generate specialized start for the current clients, we consider utilizing personalized information (few private images) to fine-tune the Stable Diffusion (SD) \citep{rombach2022high} locally, so as to obtain personalized generative model in the server. However, considering the expensive transmission cost due to the tremendous parameter weights of SD, we employ LoRA (Low-Rank Adaptation) \citep{hu2021lora} as in Figure~\ref{fig:example-image}, a parameter efficient fine-tuning strategy. LoRA freezes original model parameters and only trains matrix A and B. The update process of LoRA is $\mathbf{W}_0+\Delta \mathbf{W}=\mathbf{W}_0+\mathbf{B A}$, where $\mathbf{A, B} \in \mathbb{R}^{d \times r}$, $\mathbf{W}_0 \in \mathbb{R}^{d \times k}$ and $\mathbf{W}_0$ represents the pre-trained model weight matrix. This approach is able to maintain overall model performance without significantly increasing communication cost. For instance, each fine-tuned parameters of SD only occupies 2M, which greatly reduces the storage and transmission burdens on clients.

After SD fine-tuning stage, each client has personalized metrics A,B and prompt contained the category information (\textit{e.g.}, “A photo of airplane”). To mitigate potential privacy breaches within the prompt, each client transmits both the personalized matrix and the corresponding prompt embedding $\mathcal{P}$ to the server efficiently. We then plug different personalized metrics in Stable Diffusion models at the server to obtain personal diffusion models, and utilize prompt embedding $\mathcal{P}$ to generate corresponding synthetic data $\mathcal{S}$. We emphasize that the matrix A and B contain the client information implicitly in the compact weight spaces instead of the whole diffusion model weights, which minimizes the  transmission burden and potential privacy leakage concern. The process are described as followed:
\begin{equation}\label{generate}
    \mathcal{S}_n=\mathcal{G}(\boldsymbol{\vartheta}_n^{c}, \mathcal{P}_n^{c})
\end{equation}
where the $\mathcal{G}$ represents synthetic data generation function. $\boldsymbol{\vartheta}_n^{c}$ represents the personalized diffusion model. $c$ is the class of client $n$.

After generating synthetic set $\mathcal{S}=\left\{\left(\boldsymbol{s}_i, y_i\right)\right\}_{i=1}^{\left|\sum_{n=1}^N\mathcal{S}_n\right|}$ in the server, we train from scratch with it to initialize global model $\boldsymbol{\theta}_g$:
\begin{equation}\label{train_1}
    \boldsymbol{\theta}_g=\boldsymbol{\theta}_g-\eta \nabla \mathcal{L}_{ce}\left(\boldsymbol{\theta}_g, \boldsymbol{s}_i, y_i\right),
\end{equation}
where $\mathcal{L}_{ce}$ denotes the cross entropy loss.

\subsection{Globalization through Fine-Tuning with Synthetic Data} \label{globalization}
Warm-start establishes an advantageous commencement in FL, which exhibites superior performance across global distribution. However, the pronounced divergence between client models is still exist and may even lead to an upward trend even through pre-trained start.

In order to alleviate this, we propose a Fine-Tuning (FT) strategy with synthetic data in the server as in the Figure~\ref{fig:example-image}. Concretely, we employ synthetic data with global information to train each local model before aggregation for mitigating local model drift. Besides, In order to alleviate the degradation due to the aggregation step, we also fine-tune the aggregated model to obtain the final model. Nevertheless, the magnitude of synthetic data surpasses that of the original dataset substantially, which means employing the entirety of synthetic data for fine-tuning results in redundancy training time. To avoid it, we craft a compact synthetic dataset $\mathcal{S}_{sub}$ (20 images/class). 
\subsection{Personalization through Dynamic Self-Distillation} \label{personalization}
As in empirical study, potent personalization can be achieved by local models on the basis of the robust start. Nevertheless, it exhibits two inherent drawbacks. Firstly, Compared with random start, local fine-tuning may result in over-fitting more easily due to the robust global model and limited local data. Secondly, relying solely on the global model for personalization substantially wastes personalized knowledge, which may lead to a suboptimal initialization for personalized tasks \citep{sun2021partialfed, jin2022personalized, zhang2023fedala}. As illustrated in Figure~\ref{emp_1}b, local model discarded in the previous round exhibits higher personalized performance compared to the current global model under the pre-trained starts.

Self-distillation transfers data knowledge embedded within the teacher model to a structurally equivalent student model, contributing to the generalization ability for the student model effectively and reducing the risk of over-fitting. Similarly, we exploit the potent personalized knowledge garnered from the last round of personalized model (local model) to assist the global model fine-tuning, like in pFedSD \citep{jin2022personalized}. However, we observe that the personalized performance is unstable during these rounds, which means last personalized model may fail to represent personalized knowledge optimally as in Figure~\ref{emp_1}c. Consequently, we propose a dynamic self-distillation strategy, and we call it DSD.

\begin{algorithm}[h]
  \caption{\textsc{DSD}}
  \textbf{Input}: {Global model $\boldsymbol{\theta}_g$; Batch $B$; Training epoch E} \\
  \textbf{Output}: {Dynamic selection strategy $W_t^n$, Personal model $\boldsymbol{\theta}_p$}

  \begin{algorithmic}[1]
    \STATE \texttt{\# optimization for selection strategy}
    \FOR{each e $n\in[E]$:}
    \STATE  Updating $W_t^n$ using Equation \ref{op-for-selection};
    \STATE Composite batch parameter by:
    \STATE $I_{t_{b}}^n[i]= \begin{cases}I_{{(t-2)}_b}^n[i] & \text { if } W_b[i]=0 \\ I_{{(t-1)}_b}^n[i] & \text { if } W_b[i]=1\end{cases}$
    \ENDFOR
    \STATE \texttt{\# Personalized Self-Distillation}
    \STATE $\boldsymbol{\theta}_p = \boldsymbol{\theta}_g$
    \FOR{each e $n\in[E]$:}
    \STATE Updating $\boldsymbol{\theta}_p$ with $I_{t_{b}}^n$ using Equation \ref{self-distill};
    \ENDFOR
    
  \end{algorithmic}
\label{DSD}
\end{algorithm}

We design a learnable distilled knowledge strategy based on Kullback-Leibler (KL) divergence to select potent self-distilled knowledge. As shown in Figure~\ref{fig:example-image}, initially, we archive the logits ($I_{t-2}^n$ and $I_{t-1}^n$, where $t$ denotes the $t$-th round.) generated by personalized models of last two rounds to construct a pairwise logits pool $I_{t}^n$. Our objective is to devise a discrete dynamic selection strategy $W_t^n$, which is aimed at choosing the logit of each private image conducive to self-distillation. In order to obtain representative logits for self-distillation, we expand the predicted distribution disparity between the output logits of the global model and the selected logits. However, it is imperative that the selected logits is able to transfer data knowledge accurately. Considering the above two points, we design the following loss to optimize the dynamic selection strategy $W_t^n$:
\begin{equation}\label{op-for-selection}
    \mathcal{L}_{DSD} = -\mathcal{L}_{KL}\left(W_t^n, \boldsymbol{\theta}_g\right)+ \alpha \mathcal{L}_{CE}\left(W_t^n,\mathbf{y}_s\right),
\end{equation}
where $\mathcal{L}_{KL}$ denotes the Kullback-Leibler (KL) divergence loss. we aim to maximize $\mathcal{L}_{KL}$ to choose predictions with more diversity, while simultaneously minimizing $\mathcal{L}_{CE}$ to ensure the precision of the selected predictions.

Given the non-differentiability in discrete selection strategies, we adopt Gumbel-Softmax sampling optimization \citep{maddison2016concrete,jang2016categorical}, which is used in deep learning to approximate discrete distributions. Specifically, we define a probability selection matrix $\omega_t^c[i, j] \in[0,1]^{2 b}$ with shape $[b, 2]$, wherein each row corresponds to a logit of each image, and each column represents the probability of selecting the corresponding logit. b represents the quantity in each batch, and it is essential to ensure $\omega_t^c[i, 0] + \omega_t^c[i, 1] = 1$. The differentiable probability selection matrix with Gumbel-Softmax sampling is as followed:

\begin{equation}
    W_t^n[i, j]=\frac{\exp \left(\left(\log \omega_t^c[i, j]+G(j)\right) / \tau_t\right)}{\sum_{r \in\{0,1\}} \exp \left(\left(\log \omega_t^c[i, r]+G(r)\right) / \tau_t\right)},
\end{equation}
where $G=-\log (-\log U)$ represents the Gumbel distribution, $U$ is sampled from $U n i f(0,1)$ and j $\in$ 0,1. $\tau_t$ is the temperature, which depends the degree of discretization. The detailed optimization for selection strategy $W_t^n$ is in Algorithm \ref{DSD}.

Upon obtaining the updated $W_t^n$, we obtain the composited logits $I_{t_{new}}^n$. And we utilize it to the process of personalized self-distillation (PSD):
\begin{equation}\label{self-distill}
\mathcal{L}_{\text{PSD}}=\mathcal{L}_{CE}\left(\boldsymbol{\theta}_g,x_{i}\right)+\beta \cdot \mathcal{L}_{KL}\left(I_{t_{new}},\boldsymbol{\theta}_g(s_{i})\right),
\end{equation}
where $\mathcal{L}_{KL}$ is employed to distill the selected personal data knowledge to the global model.

\begin{table*}[ht]
\footnotesize 
\captionsetup{font=footnotesize}
\centering 
\setlength{\tabcolsep}{1mm}  
\captionof{table}{The results of global model (GM) as well as personalized model (PM). The Avg represents the average accuracy.}
\label{tab:main_result}
\vspace{-0.3cm}
\scalebox{0.9}{
\begin{tabular}{cc|ccccc|ccccccc|cccccc}
\toprule
                                                                                                             &                                       & \multicolumn{5}{c|}{Office-Caltech 10}                                                                                                                                                               & \multicolumn{7}{c|}{DomainNet}                                                                                                                                                                                                                                                        & \multicolumn{6}{c}{Camelyon17}                                                                                                                                                          \\ \cmidrule{3-20} 
\multirow{-2}{*}{Method}                                                                                     & \multirow{-2}{*}{Model}               & A                                     & C                                     & D                                    & W                                     & Avg                                   & C                                     & I                                     & P                                     & Q                                     & R                                     & S                                     & Avg                                   & H1                           & H2                           & H3                           & H4                           & H5                           & Avg                          \\ \midrule
                                                                                                             & GM                                    & 81.3                                  & 81.8                                  & 75.0                                 & 96.6                                  & 82.9                                  & 79.8                                  & 43.0                                  & 52.2                                  & 71.7                                  & 79.7                                  & 65.8                                  & 65.3                                  & 88.4                         & 86.8                         & 94.6                         & 91.4                         & 90.6                         & 90.4                         \\
\multirow{-2}{*}{FedAvg}                                                                                     & PM                                    & 87.5                                  & 81.8                                  & 75.0                                 & 98.3                                  & 85.6                                  & 81.8                                  & 50.4                                  & 60.8                                  & 91.5                                  & 84.3                                  & 73.5                                  & 73.7                                  & 89.6                         & 89.3                         & 95.4                         & 95.0                         & 93.2                         & 92.5                         \\ \midrule
                                                                                                             & GM                                    & 77.1                                  & 71.6                                  & 62.5                                 & 76.3                                  & 75.6                                  & 74.1                                  & 40.3                                  & 49.9                                  & 75.3                                  & 76.5                                  & 67.3                                  & 63.9                                  & 88.0                         & 87.8                         & 94.3                         & 91.8                         & 90.4                         & 90.8                         \\
\multirow{-2}{*}{FedBN}                                                                                      & PM                                    & 89.1                                  & 72.0                                  & 84.4                                 & 96.6                                  & 85.5                                  & 81.8                                  & 49.2                                  & 63.5                                  & 91.1                                  & 84.8                                  & 76.3                                  & 74.5                                  & 90.7                         & 90.1                         & 95.6                         & 94.6                         & 92.4                         & 92.7                         \\ \midrule
\multicolumn{2}{c|}{\textbf{Centralized}}                                                                                                            & 93.8                                  & 72.0                                  & 93.8                                 & 98.3                                  & 84.7                                  & 81.2                                  & 50.2                                  & 60.1                                  & 92.0                                  & 81.1                                  & 70.8                                  & 72.6                                  & 90.9                         & 91.5                         & 95.5                         & 95.9                         & 92.4                         & 93.3                         \\ \midrule
\multicolumn{1}{c|}{}                                                                                        & FGL                                   & 81.8                                  & 81.8                                  & 93.8                                 & 62.7                                  & 80.0                                  & 80.2                                  & 49.9                                  & 76.2                                  & 49.7                                  & 85.5                                  & 73.5                                  & 69.2                                  & 50.0                         & 50.2                         & 50.1                         & 50.1                         & 50.0                         & 50.1                         \\
\multicolumn{1}{c|}{\multirow{-2}{*}{one-shot}}                                                              & \cellcolor[HTML]{F8DEDE}\textbf{Ours} & \cellcolor[HTML]{F8DEDE}\textbf{94.8} & \cellcolor[HTML]{F8DEDE}\textbf{88.0} & \cellcolor[HTML]{F8DEDE}\textbf{100} & \cellcolor[HTML]{F8DEDE}\textbf{93.2} & \cellcolor[HTML]{F8DEDE}\textbf{94.0} & \cellcolor[HTML]{F8DEDE}\textbf{78.7} & \cellcolor[HTML]{F8DEDE}\textbf{58.3} & \cellcolor[HTML]{F8DEDE}\textbf{77.0} & \cellcolor[HTML]{F8DEDE}\textbf{84.0} & \cellcolor[HTML]{F8DEDE}\textbf{84.3} & \cellcolor[HTML]{F8DEDE}\textbf{76.7} & \cellcolor[HTML]{F8DEDE}\textbf{76.5} & \cellcolor[HTML]{F8DEDE}78.8 & \cellcolor[HTML]{F8DEDE}73.0 & \cellcolor[HTML]{F8DEDE}85.1 & \cellcolor[HTML]{F8DEDE}79.7 & \cellcolor[HTML]{F8DEDE}73.9 & \cellcolor[HTML]{F8DEDE}78.1 \\ \midrule
                                                                                                             & GM                                    & 88.5                                  & 87.1                                  & 93.8                                 & 84.8                                  & 88.5                                  & 89.3                                  & 60.0                                  & 82.0                                  & 74.7                                  & 92.7                                  & 86.3                                  & 80.8                                  & 87.8                         & 84.7                         & 88.0                         & 89.5                         & 86.1                         & 87.2                         \\
\multirow{-2}{*}{\begin{tabular}[c]{@{}c@{}}Pre-IN1K\\ (5-rounds)\end{tabular}}                              & PM                                    & 94.8                                  & 72.0                                  & 93.8                                 & 96.6                                  & 89.3                                  & 93.1                                  & 67.8                                  & 86.4                                  & 92.7                                  & 94.9                                  & 88.4                                  & 87.2                                  & 90.8                         & 90.0                         & 94.3                         & 95.0                         & 93.0                         & 92.6                         \\
                                                                                                             & GM                                    & 94.8                                  & 88.9                                  & 90.6                                 & 88.1                                  & 90.6                                  & 88.4                                  & 60.3                                  & 78.2                                  & 84.0                                  & 89.1                                  & 86.9                                  & 81.2                                  & 85.0                         & 80.9                         & 90.2                         & 85.4                         & 89.4                         & 86.2                         \\
\multirow{-2}{*}{\begin{tabular}[c]{@{}c@{}}FGL\\ (5-rounds)\end{tabular}}                                   & PM                                    & 94.3                                  & 28.4                                  & 96.9                                 & 91.5                                  & 77.8                                  & 89.1                                  & 65.4                                  & 79.1                                  & 92.8                                  & 92.7                                  & 86.6                                  & 84.3                                  & 87.5                         & 85.9                         & 94.5                         & 94.3                         & 93.2                         & 91.1                         \\
\rowcolor[HTML]{F8DEDE} 
\cellcolor[HTML]{F8DEDE}                                                                                     & \textbf{GM}                           & \textbf{93.8}                         & \textbf{90.7}                         & \textbf{100}                         & \textbf{100}                          & \textbf{96.1}                         & \textbf{91.5}                         & \textbf{66.0}                         & \textbf{83.0}                         & \textbf{88.3}                         & \textbf{90.9}                         & \textbf{88.8}                         & \textbf{84.8}                         & 91.0                         & 89.4                         & 91.4                         & 92.4                         & 89.3                         & 90.7                         \\
\rowcolor[HTML]{F8DEDE} 
\multirow{-2}{*}{\cellcolor[HTML]{F8DEDE}\textbf{\begin{tabular}[c]{@{}c@{}}Ours\\ (5-rounds)\end{tabular}}} & {\color[HTML]{000000} \textbf{PM}}    & {\color[HTML]{000000} \textbf{94.3}}  & {\color[HTML]{000000} \textbf{90.7}}  & {\color[HTML]{000000} \textbf{100}}  & {\color[HTML]{000000} \textbf{100}}   & {\color[HTML]{000000} \textbf{96.2}}  & {\color[HTML]{000000} \textbf{93.1}}  & {\color[HTML]{000000} \textbf{69.5}}  & {\color[HTML]{000000} \textbf{87.7}}  & {\color[HTML]{000000} \textbf{93.6}}  & {\color[HTML]{000000} \textbf{94.2}}  & {\color[HTML]{000000} \textbf{90.3}}  & {\color[HTML]{000000} \textbf{88.1}}  & \textbf{91.3}                & \textbf{92.4}                & \textbf{95.9}                & \textbf{95.6}                & \textbf{93.6}                & \textbf{94.4}                \\ \bottomrule
\end{tabular}}
\end{table*}

\begin{table*}[t]
\footnotesize  
\captionsetup{font=footnotesize}
\vspace{-0.3cm}
\centering 
\setlength{\tabcolsep}{1mm}  
\captionof{table}{The results of global performance on unseen clients.}
\label{unseen}
\vspace{-0.3cm}
\scalebox{0.9}{
\begin{tabular}{c|ccccc|ccccccc|cccccc}
\toprule
                          & \multicolumn{5}{c|}{Office-Caltech 10 (unseen)}                                & \multicolumn{7}{c|}{DomainNet (unseen)}                                                                        & \multicolumn{6}{c}{Camelyon17 (unseen)}                                                       \\ \cmidrule{2-19} 
\multirow{-2}{*}{Dataset} & A             & C             & D             & W             & Avg           & C             & I             & P             & Q             & R             & S             & Avg           & h1            & h2            & h3            & h4            & h5            & Avg           \\ \midrule
Centralized               & 93.8          & 72.0          & 93.8          & 98.3          & 84.7          & 81.2          & 50.2          & 60.1          & 92.0          & 81.1          & 70.8          & 72.6          & 90.9          & 91.5          & 95.5          & 95.9          & 92.4          & 93.3          \\
FedAvg                    & 69.3          & 43.1          & 68.8          & 79.7          & 65.2          & 65.2          & 34.4          & 42.7          & 37.7          & 63.2          & 44.0          & 47.9          & 84.5          & 78.1          & 84.1          & 85.0          & 81.7          & 82.8          \\ \midrule
Pre-IN1K                  & 89.1          & 85.8          & 87.5          & 78.0          & 85.1          & 86.8          & 49.0          & 73.2          & 54.6          & 87.8          & 75.9          & 71.2          & 83.9          & 80.8          & 83.5          & 84.1          & 83.9          & 83.2          \\
FGL                       & 91.2          & 81.8          & 93.8          & 77.9          & 86.2          & 84.2          & 48.8          & 72.1          & 54.0          & 87.2          & 72.7          & 72.3          & 82.7          & 79.5          & 83.9          & 83.4          & 83.3          & 82.6          \\
\rowcolor[HTML]{F8DEDE} 
\textbf{Ours}             & \textbf{91.7} & \textbf{88.0} & \textbf{96.9} & \textbf{98.3} & \textbf{93.7} & \textbf{90.3} & \textbf{56.9} & \textbf{77.8} & \textbf{80.5} & \textbf{87.7} & \textbf{81.2} & \textbf{79.1} & \textbf{88.0} & \textbf{87.2} & \textbf{84.1} & \textbf{91.1} & \textbf{84.1} & \textbf{86.9} \\ \bottomrule
\end{tabular}}
\end{table*}

\section{Experiment}\label{sec:exp}
To better illustrate the effectiveness of our method, we evaluate the efficacy of both global and personalized models under \textbf{cross-domain shift} within FL. We further present the efficacy of our approach under \textbf{label skew} in the \textbf{Appendix}. We leverage two publicly available benchmark datasets of natural scene (\textbf{DomainNet} \citep{peng2019moment}, \textbf{Office-Caltech10} \citep{gong2012geodesic}) and real-word medical datasets (\textbf{Camelyon17}) \citep{chen2023federated}. The details for the datasets are in \textbf{Appendix}.

\subsection{Baselines} 
To illustrate the performance of both global and personalized model, we conduct a comparative study in framworks with random start (FedAvg \citep{konevcny2016federated}, FedBN\citep{li2021fedbn}) and pre-trained starts (Pre-ImageNet1k (Pre-IN1K) \citep{nguyen2022begin}, FGL \citep{zhang2023federated}). As FedBN focuses on personalization, we obtain the global model by aggregating local models at server. And other methods only target on the globalization, we evaluate their ability on the personalized model via local model. To compare the performance of the pre-trained start, our study also undertakes a comparative analysis of the performance of one-shot communication exhibited by WarmFed and FGL. Moreover, to emphasize the efficiency brought by pre-trained start, we compare with pre-trained starts over just 5 communication rounds with that of FedAvg over 200 communication rounds. The more introduction about experiment settings and details are in \textbf{Appendix}.

\subsection{Main Results}

\textbf{Results on Office-Caltech 10 and DomainNet.} \textit{WarmFed demonstrates leading proficiency across one-shot and 5-rounds communication(s).} As presented in Table~\ref{tab:main_result}, FL with pretrained starts outperforms random start in both global and personalized model.  Specially, WarmFed exhibits unparalleled performance in both one-shot and 5-rounds. \textit{Initially, with respect to the one-shot communication}, WarmFed attains 94.00\% and 76.51\% on Office-Caltech 10 and DomainNet, outperforming centralized learning and FGL, respectively. \textit{Secondly, with respect to 5-rounds}, we observe that both global and personalized models exhibit sustained advancement. For example, the global performances of Office-Caltech 10 and DomainNet achieve 96.11\% and 84.75\% impressively, which surpasses other frameworks with random or pre-trained start.  

The trend of personal model is consistent with that of the global model. However, we find that the overall personalized effect of Pre-IN1k remains superior to FGL, which suggests that despite FGL offering clients a robust start, the inherent biases present in synthetic data lead to accumulated client drift in early communications.

\textbf{Results on Camelyon17.} \textit{WarmFed also showcases strong adaptability to intricate medical dataset.}  We only employ synthetic data twice as much as the private data to train DenseNet121, which achieves 78.06\% in one-shot as in Table~\ref{tab:main_result}. Conversely, FGL with the same number of synthetic set only achieves 50.08\%. This demonstrates that when dealing with intricate data, the limitations of generating synthetic data in FGL are accentuated. And it is observed that the overall performance of FGL also falls short of that achieved by pre-ImageNet. However, our global and personal models achieve better results compared with random start and surpass both FGL and pre-ImageNet in 5 rounds. 

\subsection{Discussions on Warm-Start}\label{key-factor}
\begin{figure}[t]
    \vspace{-0.5cm}
    \centering
    \includegraphics[width=0.8\columnwidth]{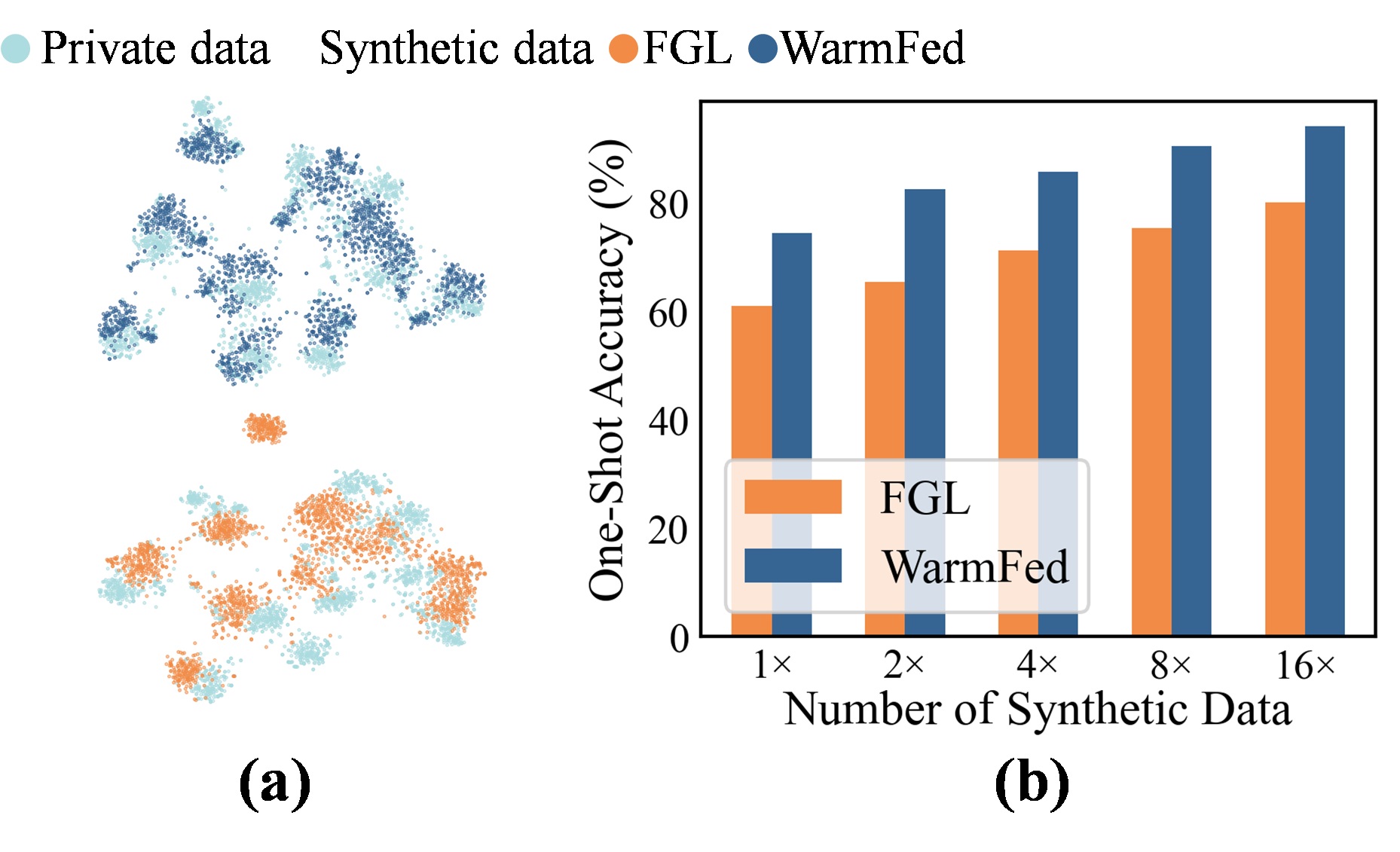}
    \vspace{-0.4cm}
    \caption{\small \textbf{(a)} The feature distribution comparison of synthetic data with private data (Office-Caltech 10). \textbf{(b)} One-shot performance under different amounts of synthetic data relative to the amount of private data (Office-Caltech 10). 
    }
    \label{fig:num_all}
    \vspace{-0.6cm}
\end{figure}

\textbf{The distribution of synthetic data.}  Given that synthetic data directly influences the effect of warm-start, we compare the feature distributions by employing the identical quantity of synthetic data and private data. As depicted in the Figure~\ref{fig:num_all}a, we observe that distribution of our synthetic data  exhibits a better fit to the real data distribution compared to FGL. This demonstrates that WarmFed is capable of delivering high-quality synthetic data for achieving a solid start for FL more easily.

\textbf{The number of synthetic data.} As indicated in Table \ref{tab:main_result}, Warm Start (one-shot) exhibit a considerable improvement compared to FGL and even centralized learning. However, considering the stochastic number of synthetic set and advantages of the device in the server, we conduct experiments for the quantity of synthetic data at various multiples relative to the real set of DomainNet. As illustrated in Figure~\ref{fig:num_all}b, there is a stable increasing in the performance of the global model as the escalating quantity of synthetic set. Besides, WarmFed consistently outperforms FGL across different orders of magnitude. This observation illustrates that server can adapt synthetic data to different scales according to its storage capacity, ensuring the creation of a robust start.

\textbf{Domain generalization.} We find that warm-start can improve generalization performance in FL, that is, the performance of the global model on unseen clients. It is a more practical setting compared with clients within static scenarios. To verify this, we designate each domain as an unseen client and train global model based on other clients. The performance of global model on each unseen client is illustrated in Table ~\ref{unseen}.  As shown in Table~\ref{tab:main_result} and Table~\ref{unseen}, client with participation or non-participation significantly impacts the final performance. However, the efficacy of pre-trained starts have achieved a consistent enhancement compared with FedAvg for Office-Caltech 10 and DomainNet. For Camelyon17, compared to FedAvg, FGL and Pre-IN1K fail to exhibit notable advancements. Conversely, WarmFed still demonstrates effective generalization ability on it. These experimental results reveal that warm-start can cope well with domain generalization in FL.

\begin{figure}[t]
    \vspace{-0.3cm}
    \centering
    \centering
    \includegraphics[width=0.6\columnwidth]{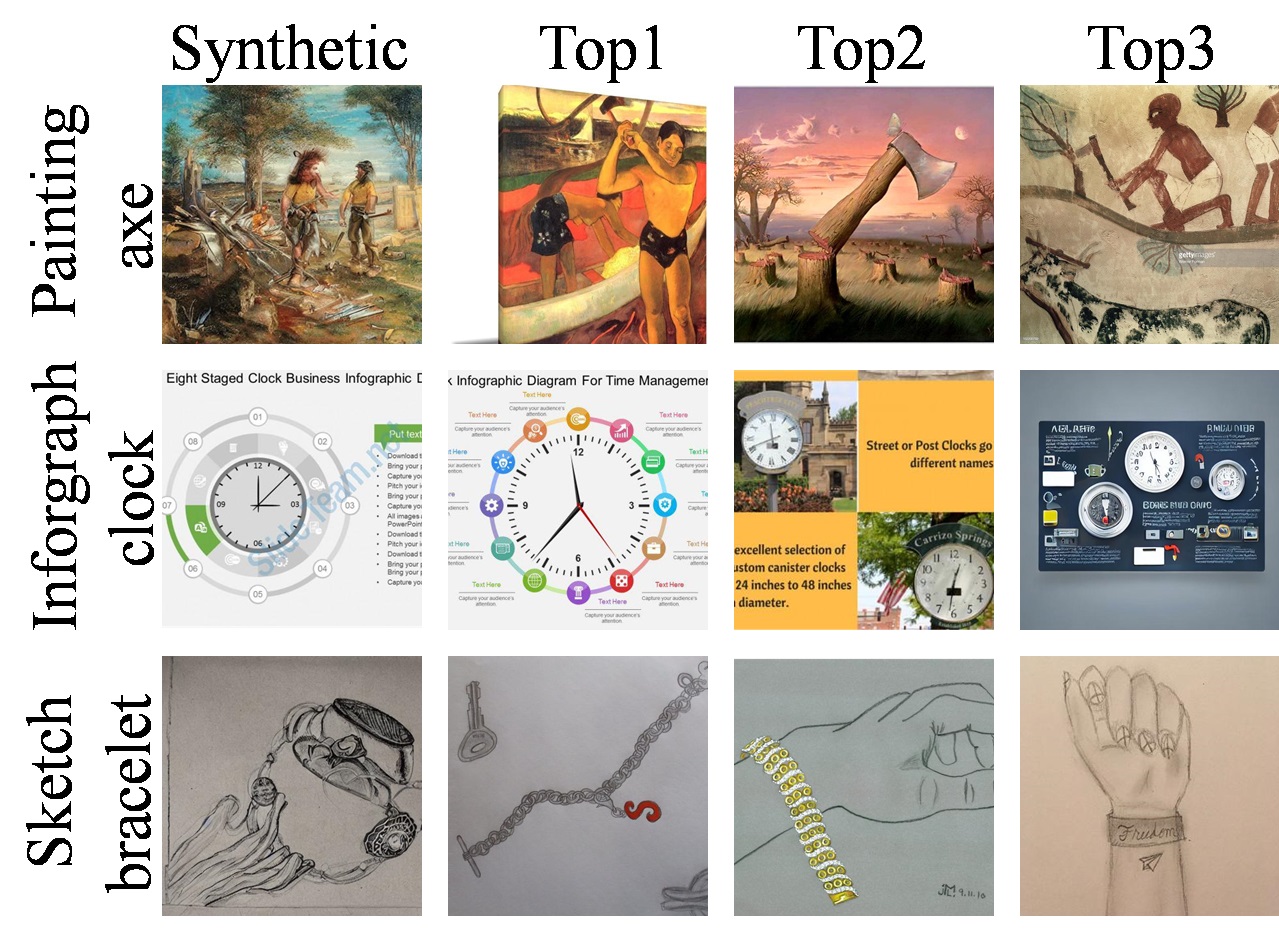}
    \vspace{-0.4cm}
    \caption{\small Visualization on synthetic and retrieved real data.}
    \label{fig:emp}
    \vspace{-0.8cm}
\end{figure}

\textbf{The privacy concerns.} Warm-Start focuses on transmitting the parameter matrix, which substantially mitigates risk of privacy leakage. Nonetheless, the synthetic data may introduce privacy concerns for client data. To substantiate the security of WarmFed, we follow the previous works \cite{somepalli2023diffusion} and detect any instances of content duplication between synthetic image and private images. We rank the similarity and present the top 3 images with their corresponding real data in Figure. The results clearly indicate that these images lack meaningful similarities in both background and foreground, confirming that WarmFed does not compromise the privacy of private data. Due to space limitations, we provide further analysis and experiments on privacy in \textbf{Appendix}.

\subsection{Ablation Study} \label{ablation_all}
we take a comprehensive investigation into the effect of server-side fine-tuning (FT) and client-side self-distillation (DSD), focusing on their impact on global and personalized models, respectively. Our experiments are conducted across Office-Caltech 10, DomainNet, and Camelyon17.

\textbf{Server-Side Fine-Tuning.} We study the effect of the server-side fine-tuning (FT) on globalization and show the results in Table~\ref{ablation_1}. It is observed that the adoption of FT results in performance gains in the global model. Specifically, compared with employing solely the warm-start strategy, there is 1\% and 2\% enhancement observed for the Domainnet and Camelyon17. This result indicates that FT effectively enhances the effect of global information, with substantial impact observed in more complex datasets.

\begin{figure}[t]
    \centering
    \includegraphics[width=0.9\columnwidth]{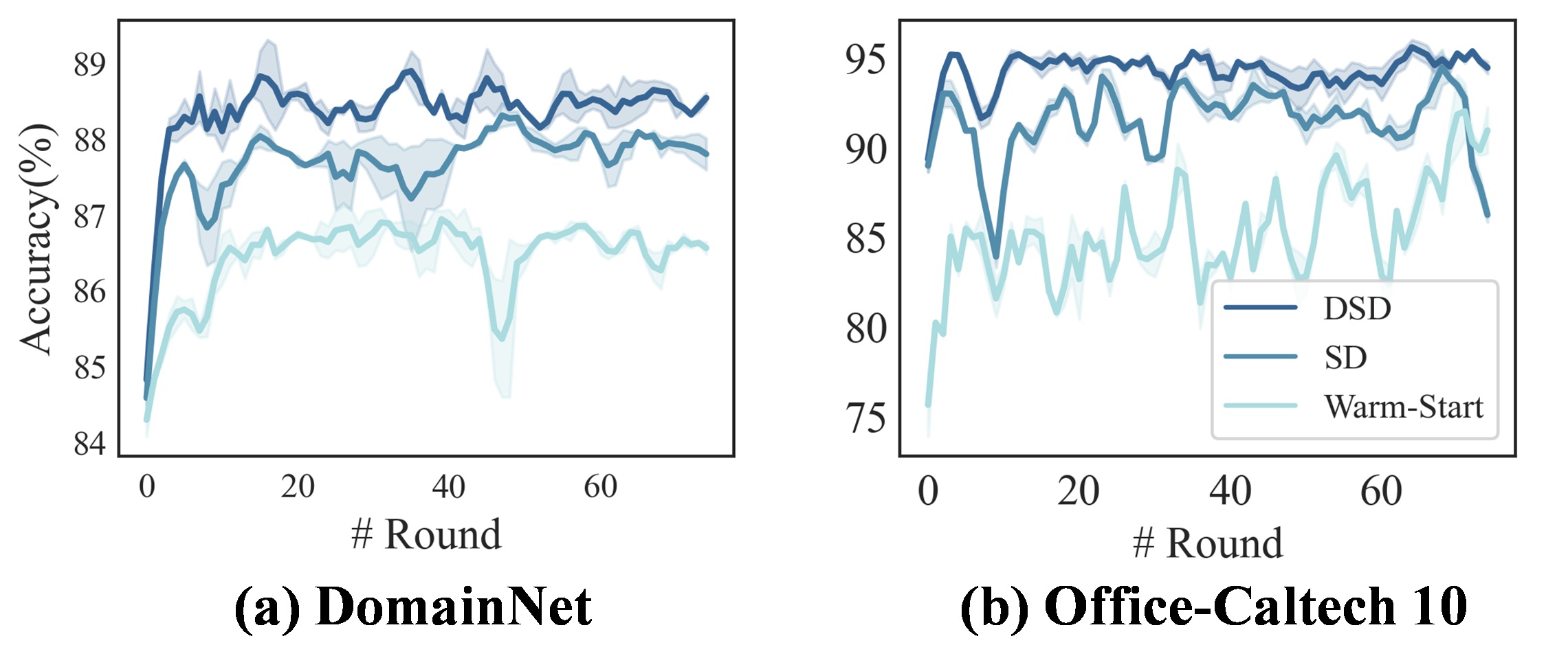}
    \vspace{-0.4cm}
    \caption{\small The performance comparison of personalized model among DSD, pFedSD (SD), Warm-Start.}
    \label{ablation_2}
    \vspace{-0.5cm}
\end{figure}

\begin{table}[h]
\footnotesize  
\captionsetup{font=footnotesize}
\centering 
\vspace{-0.2cm}
\setlength{\tabcolsep}{1mm}  
\captionof{table}{The results of ablation results of FT and DSD.}
\label{ablation_1}
\vspace{-0.3cm}
\begin{tabular}{cccc}
\toprule
FT                      & Office-Caltech 10 & DomainNet      & Camelyon17     \\ \midrule
                        & 95.5              & 83.18          & 88.78          \\
\textbf{\checkmark
}              & \textbf{96.11}    & \textbf{84.75} & \textbf{90.71} \\ \midrule
\multicolumn{1}{l}{DSD} & Office-Caltech 10 & DomainNet      & Camelyon17     \\
                        & 90.4              & 86.5           & 92.2           \\
\textbf{\checkmark
}              & \textbf{96.2}     & \textbf{88.1}  & \textbf{94.1}  \\ \bottomrule
\end{tabular}
\vspace{-0.3cm}
\end{table}

\textbf{Client-Side Dynamic Self-Distillation.} We study the effect of the dynamic self-distillation (DSD) on personalization. As illustrated in Table \ref{ablation_1}, DSD leads to solid gains. To further elucidate the superiority of DSD, we present a comparison across 75 communication rounds. As depicted in Figure~\ref{ablation_2}, although the personalized performance of local model (warm-start) yields a discernible level of personalization, the degree of it fluctuates significantly, notably evidenced in the Office-Caltech 10. This immensely exerts a significant influence on the process of personalization, and introduces uncertainty to SD (pFedSD). However, it is evident that DSD consistently outperforms SD, which further underscores the capacity of dynamic selection process in DSD to furnish richer personalized knowledge for the distillation process.

\subsection{Complexity Analysis}
\textbf{Communication cost.} WarmFed requires the transmission of additional matrices, incurring a minor, one-time cost. However, as the number of client grows, this transmission cost also scales up. To explore transmission efficiency of WarmFed, we give experiments on DomainNet with 60 clients (10 times). We start from two different perspective: 1) the globalized performance under consistent communication costs, and 2) the communication cost incurred for achieving convergence. The cumulative size of model parameters (include the extra cost in FGL and WarmFed) encompassing all clients are adopted as the total communication cost of each round.

\begin{figure}[t]
    \centering
    \includegraphics[width=0.9\columnwidth]{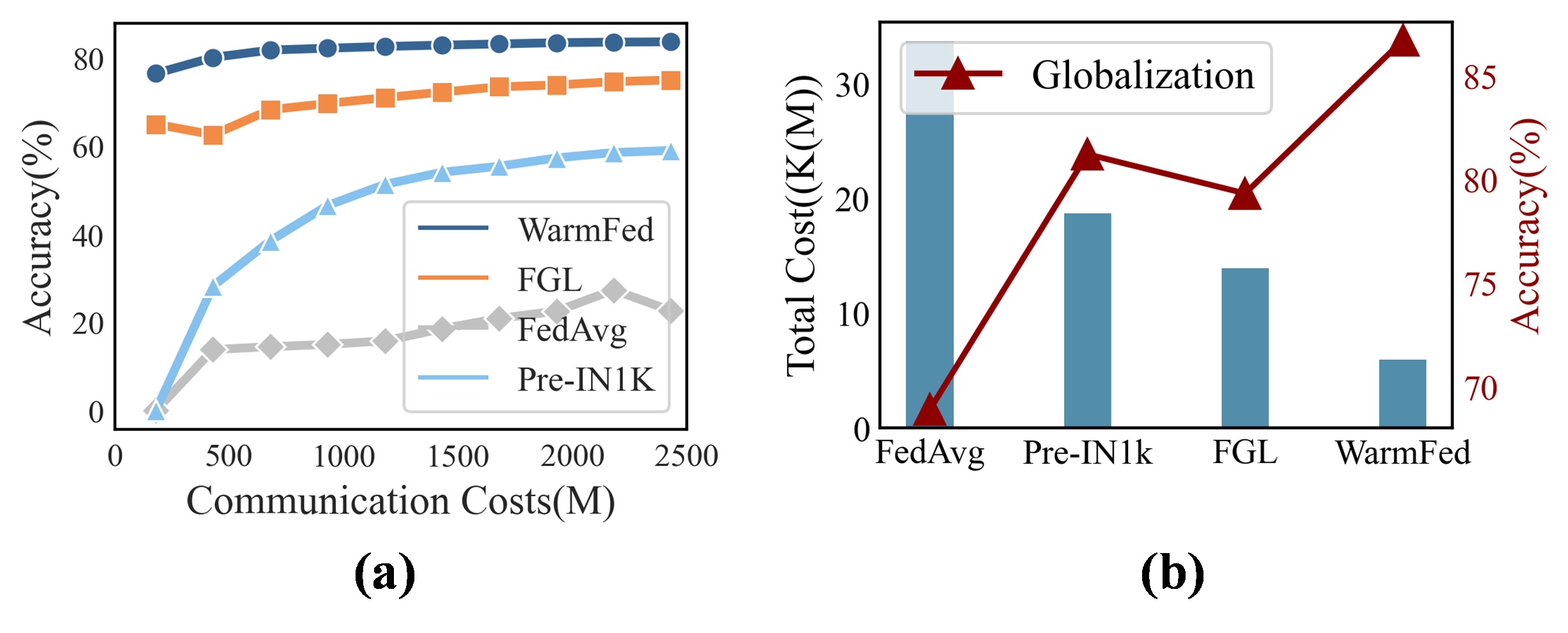}
    \vspace{-0.4cm}
    \caption{\small The cost in clients $\times$ 10 on DomainNet. \textbf{(a)} The performance within the uniform communication costs. \textbf{(b)} Communication costs required for convergence.}
    \label{cost}
    \vspace{-0.3cm}
\end{figure}

As in Figure \ref{cost}a, WarmFed outperforms FedAvg and other pre-trained methods. Furthermore,  Figure \ref{cost}b reveals that WarmFed exhibits the lowest communication cost until achieving convergence, while gives a better performance of globalization. These observations have demonstrated that the extra matrix parameters conveyed in the first round of WarmFed do not compromise the overarching efficiency even under considerable clients.

\begin{table}[h]
\footnotesize  
\captionsetup{font=footnotesize}
\centering 
\setlength{\tabcolsep}{1mm}  
\captionof{table}{Average Time Per Round on Office-Caltech 10, and time on the Server, Client, and computational cost per client}
\label{Computation}
\vspace{-0.3cm}
\begin{tabular}{cccccc}
\toprule
FT         & DSD        & \begin{tabular}[c]{@{}c@{}}Times\\ (s)\end{tabular} & \begin{tabular}[c]{@{}c@{}}Server\\ (s)\end{tabular} & \begin{tabular}[c]{@{}c@{}}Clients\\ (Avg/s)\end{tabular} & \begin{tabular}[c]{@{}c@{}}Computational Cost\\ (G)\end{tabular} \\ \midrule
           &            & 33.8                                                & 0.03                                                 & 6.9                                                       & 835.05                                                                     \\
\textbf{\checkmark} & \textbf{}  & 45.4                                                & 12.8(6.1)                                                  & 7.0                                                       & 835.05                                                                     \\
\textbf{}  & \textbf{\checkmark} & 58.36                                               & 0.03                                                 & 12.8                                                      & 835.05/898.08                                                              \\ \bottomrule
\end{tabular}
\vspace{-0.4cm}
\end{table}

\textbf{Computation complexity. }WarmFed includes both globalization and personalization stages into the subsequent communications. For globalization, since the server is not limited by device, we don't consider the computational complexity in FT, while it might lengthen whole time. Additionally, DSD leads to increased computational complexity and processing time in the client. To explore the computational complexity of WarmFed, we compared the execution time (include communication, the time in the server and client), and the number of floating-point operations (FLOPs) of each client as a measure of computational efficiency.

As indicated in table~\ref{Computation}, WarmFed achieves the shortest execution time without FT and DSD. Firstly, FT does lead to a time increase in the server. But this can be mitigated by employing parallel algorithms on the server according to the device. For example, we observe that parallel execution with 2 GPUs reduce the execution time to 6.1s. Utilizing more GPUs could further decrease this time, which makes it more efficient. Secondly, DSD adds approximately 6s to the client’s processing time. We also calculate the client’s FLOPs per round. Since DSD selects informative logits and aligns them for self-distillation, logits can be obtained either by retaining the previous personalized models or directly retaining all logits. The results show that DSD imposes no extra computational cost when using the first approach, while the second approach increases computational costs by 7.5\%. Nevertheless, the modest increase in time and computational cost yields a stable improvement in personalization as in Figure~\ref{ablation_2}.

\section{Conclusion}
In this paper, we start from the initialization of FL,  concurrently acquiring both potent global and personalized models. We propose WarmFed, an innovative FL framework. WarmFed achieves warm-start by efficiently transmitting the parameter matrix from clients without privacy leakage. Based on this, we propose a client-side fine-tuning and dynamic self-distillation strategy to achieve better globalization and personalization. Extensive experimentation validates the  considerable performance of WarmFed in only one and five round(s).

\bibliography{aaai25}

\clearpage

\section*{AAAI Paper Checklist}

\paragraph{This paper:}
\begin{itemize}
\item Includes a conceptual outline and/or pseudocode description of AI methods introduced. [yes]
\item Clearly delineates statements that are opinions, hypothesis, and speculation from objective facts and results. [yes]
\item Provides well marked pedagogical references for less-familiare readers to gain background necessary to replicate the paper. [yes]

\end{itemize}

\paragraph{Does this paper make theoretical contributions?} [no]
\begin{itemize}
\item All assumptions and restrictions are stated clearly and formally. [yes]
\item All novel claims are stated formally (e.g., in theorem statements). [yes]
\item Proofs of all novel claims are included. [no]
\item Proof sketches or intuitions are given for complex and/or novel results. [yes]
\item Appropriate citations to theoretical tools used are given.[yes]
\item All theoretical claims are demonstrated empirically to hold. [yes]
\item All experimental code used to eliminate or disprove claims is included. [yes]
\end{itemize}

\paragraph{Does this paper rely on one or more datasets?} [yes]
\begin{itemize}
    \item A motivation is given for why the experiments are conducted on the selected datasets. [yes]
\item All novel datasets introduced in this paper are included in a data appendix. [yes]
\item All novel datasets introduced in this paper will be made publicly available upon publication of the paper with a license that allows free usage for research purposes.[yes]
\item All datasets drawn from the existing literature (potentially including authors’ own previously published work) are accompanied by appropriate citations. [yes]
\item All datasets drawn from the existing literature (potentially including authors’ own previously published work) are publicly available. [yes]
\item All datasets that are not publicly available are described in detail, with explanation why publicly available alternatives are not scientifically satisficing. [yes]
\end{itemize}

\paragraph{Does this paper include computational experiments?} [yes]

\begin{itemize}
    \item Any code required for pre-processing data is included in the appendix. [yes]
\item All source code required for conducting and analyzing the experiments is included in a code appendix. [yes]
\item All source code required for conducting and analyzing the experiments will be made publicly available upon publication of the paper with a license that allows free usage for research purposes. [yes]
\item All source code implementing new methods have comments detailing the implementation, with references to the paper where each step comes from [yes]
\item If an algorithm depends on randomness, then the method used for setting seeds is described in a way sufficient to allow replication of results. [yes]
\item This paper specifies the computing infrastructure used for running experiments (hardware and software), including GPU/CPU models; amount of memory; operating system; names and versions of relevant software libraries and frameworks.[yes]
\item This paper formally describes evaluation metrics used and explains the motivation for choosing these metrics. [yes]
\item This paper states the number of algorithm runs used to compute each reported result. [yes]
\item Analysis of experiments goes beyond single-dimensional summaries of performance (e.g., average; median) to include measures of variation, confidence, or other distributional information.[yes]
\item The significance of any improvement or decrease in performance is judged using appropriate statistical tests (e.g., Wilcoxon signed-rank). [yes]
\item This paper lists all final (hyper-)parameters used for each model/algorithm in the paper’s experiments. {yes}
\item This paper states the number and range of values tried per (hyper-) parameter during development of the paper, along with the criterion used for selecting the final parameter setting. [yes]
\end{itemize}

\end{document}